\documentclass[letterpaper, 10 pt, conference]{ieeeconf}

\usepackage{xcolor}
\usepackage{tabularx}

\IEEEoverridecommandlockouts                              
\usepackage[T1]{fontenc}

\usepackage[noadjust]{cite}

\usepackage{filecontents}
\usepackage{amsmath}
\DeclareMathSymbol{:}{\mathord}{operators}{"3A}  
\usepackage{yhmath} 
\usepackage{mathtools} 
\usepackage{amssymb} 
\usepackage{xfrac} 
\usepackage{sidecap}
\usepackage{graphicx}
\usepackage{subcaption}
\usepackage{multirow} 
\usepackage{fancyhdr}
\usepackage{leftidx} 
\usepackage{cancel} 
\usepackage{caption}
\usepackage[super]{nth} 
\usepackage[normalem]{ulem} 
\usepackage{color}
\usepackage{xcolor} 
\usepackage{booktabs} 
\usepackage{siunitx} 
\usepackage{booktabs} 
\usepackage{colortbl} 
\usepackage{textcomp} 
\usepackage{tikz}
\usepackage[noadjust]{cite} 
\usepackage{xcolor}
\usepackage{wasysym}  

\newcommand{\RNum}[1]{\uppercase\expandafter{\romannumeral #1\relax}} 

\newcommand{\realfield}[1]{\hbox{I \kern -.5em R}^{#1}}
\newcommand {\mb}[1]{\mathbf{#1}}

\newcommand{\uvec}[1]{\hat{\mathbf{#1}}}

\newcommand{\T}{^{\mathrm{T}}}  

\usepackage{textcomp}
\let\dollar\textdollaroldstyle

\makeatletter
\newcommand{\ostar}{\mathbin{\mathpalette\make@circled\star}}
\newcommand{\make@circled}[2]{%
  \ooalign{$\m@th#1\smallbigcirc{#1}$\cr\hidewidth$\m@th#1#2$\hidewidth\cr}%
}
\newcommand{\smallbigcirc}[1]{%
  \vcenter{\hbox{\scalebox{0.77778}{$\m@th#1\bigcirc$}}}%
}
\makeatother

\usepackage{enumerate}

\usepackage{color} 

\usepackage[textsize=scriptsize, bordercolor=white,backgroundcolor=gray!30,linecolor=black,colorinlistoftodos]{todonotes}  
\usepackage[normalem]{ulem} 
\usepackage{soul} 

\usepackage{algorithm2e}
\usepackage{algpseudocode}
\RestyleAlgo{ruled}

\RestyleAlgo{ruled}

\algnewcommand\algorithmicinput{\textbf{Input:}}
\algnewcommand\Input{\item[\algorithmicinput]}

\algnewcommand\algorithmicoutput{\textbf{Output:}}
\algnewcommand\Output{\item[\algorithmicoutput]} 

\graphicspath{{figures/pdf/}}

\newbox\tempbox
\makeatletter
\let\NAT@parse\undefined
\makeatother
\usepackage{hyperref} 
\hypersetup{
  colorlinks   = true, 
  urlcolor     = blue, 
  linkcolor    = blue, 
  citecolor   = green 
}

\usepackage{balance}
\title{\LARGE \bf Multi-Modal Gesture Recognition from Video and\\ Surgical Tool Pose Information via Motion Invariants}
\author{Jumanh Atoum$^{1}$, Garrison L.H. Johnston$^{2}$, Nabil~Simaan$^{2}$, and Jie Ying Wu$^{1}$
\thanks{$^{1}$Department of Computer Science, Vanderbilt University, Nashville, TN 37212, USA
        {\tt\small (Jumanh.Atoum, JieYing.Wu) @vanderbilt.edu}}%
\thanks{$^{2}$Department of Mechanical Engineering, Vanderbilt University, Nashville, TN 37235, USA
        {\tt\small (Garrison.L.Johnston, Nabil.Simaan) @vanderbilt.edu}}
\thanks{This work was supported through the Wellcome Leap SAVE Program and Vanderbilt University internal funds.}%
}

\begin{document}

\maketitle
\thispagestyle{empty}  


\thispagestyle{fancy}
\fancyhf{}
\renewcommand{\headrulewidth}{0pt}
\lhead{} 
\rfoot{\centering \scriptsize \copyright 2024 IEEE. Personal use of this material is permitted. Permission from IEEE must be obtained for all other uses, in any current or future media, including reprinting/republishing this material for advertising or promotional purposes, creating new collective works, for resale or redistribution to servers or lists, or reuse of any copyrighted component of this work in other works.}

\pagestyle{empty}


\begin{abstract}

Recognizing surgical gestures in real-time is a stepping stone towards automated activity recognition,  skill assessment, intra-operative assistance, and eventually surgical automation. The current robotic surgical systems provide us with rich multi-modal data such as video and kinematics. While some recent works in multi-modal neural networks learn the relationships between vision and kinematics data, current approaches treat kinematics information as independent signals, with no underlying relation between tool-tip poses. However, instrument poses are geometrically related, and the underlying geometry can aid neural networks in learning gesture representation. Therefore, we propose combining motion invariant measures (curvature and torsion) with vision and kinematics data using a relational graph network to capture the underlying relations between different data streams. We show that gesture recognition improves when combining invariant signals with tool position, achieving 90.3\% frame-wise accuracy on the JIGSAWS suturing dataset. Our results show that motion invariant signals coupled with position are better representations of gesture motion compared to traditional position and quaternion representations. Our results highlight the need for geometric-aware modeling of kinematics for gesture recognition.







\end{abstract}


\section{Introduction} \label{sec:intro}

\par Robot-assisted surgery has played an important role in improving minimally invasive surgeries~\cite{probst2023review}. Not only do robots enable more precise controls and better ergonomics for the surgeon compared with traditional laparoscopic surgery, but they also capture multi-modal information that could inform computational models of gestures~\cite{guthart2000intuitive}. 
Models of surgical gestures can enable downstream tasks like skill assessment and task automation. To improve the generalizability of these gestures, works have been proposed to deconstruct long sequences into atomic gestures~\cite{tao2012sparse, long2021relational, lea2016temporal, wu2021cross, van2020multi, kiyasseh2023vision, van2021gesture, qin2020temporal, hutchinson2023towards, geist2024learning}. In laparoscopic surgical robotic systems, the endoscopic camera feed provides the vision information while the robotic arms provide the kinematics data containing tool-tip pose information~\cite{gao2014jhu}. To date, most works encode kinematic signals from tool-tip pose or velocity without considering geometric constraints informing their relationships~\cite{weerasinghe2024multimodal, long2021relational, chen2005formulas}. This requires the machine learning models to learn the underlying geometry of tool-tip motion. The geometry of tool-tip motion can be understood in part by using \emph{motion invariants}. These parameters uniquely characterize motion as they remain constant (i.e. invariant) regardless of the choice of coordinate system. In this paper, we will use the tool-tip's \emph{curvature} and \emph{torsion}, which measure the closeness of the motion to a straight line and to a planar curve, respectively. We hypothesize that directly computing motion invariants from tool-tip trajectories~\cite{boutin2000numerically,calabi1998differential} and using them as inputs to machine-learning models can improve gesture recognition.

Real-time surgical gesture recognition is challenging because of frequent state transitions, disturbances in sensory data, mechanical manipulation patterns, and varying proficiency of the surgeons~\cite{long2021relational,liu2021towards}. To address these challenges, many approaches have been studied and developed in the past decade~\cite{zappella2013surgical, tao2012sparse, gurcan2019surgical, van2020multi}. One of the early approaches modeled sequential robotic kinematics data (e.g., the pose and velocity of the tool-tips) with traditional machine learning methods such as linear classifiers with customized metrics~\cite{zappella2013surgical} and variants of hidden Markov models~\cite{tao2012sparse}. 
More recent approach~\cite{shi2022recognition} used kinematics data for gesture recognition, with a transformer model, achieving frame-wise accuracy of 89.3\%. Other works~\cite{li2023robotic, gurcan2019surgical, agarwal2022temporal, hutchinson2023towards} focused on improving visual representations. Gurcan et al.~\cite{gurcan2019surgical} used synthetic videos to augment the JIGSAWS dataset to train a multi-scale recurrent network for endoscopic video and achieved a 90.2\% frame-wise accuracy. While the accuracy is promising, the augmentations employed assumed a specific setup and could lead the model to overfit on JIGSAWS. Moreover, these unimodal methods do not account for interactions between the multi-modal signals. 

The interaction between motion trajectories and vision data play an important role in gesture prediction, gesture recognition, and skill assessment~\cite{long2021relational, weerasinghe2024multimodal, shi2022recognition, qin2020temporal, wu2021cross}. These works use multi-modal networks to model the interactions between the kinematics data (position and rotation) and visual data (optical flow or RGB images) for gesture recognition tasks. However, using tool-tip pose and/ or velocity is insufficient to describe geometric shapes from surgical gestures. Additionally, surgical datasets are generally small, which limits training data for neural networks. Thus, we hypothesize that using motion-invariant representations from trajectories as input would improve neural networks' training efficiency.

To better describe trajectories and shapes, previous works~\cite{boutin2000numerically, calabi1998differential} used differential geometry methods to extract 3-D invariant signals (e.g., curvature and torsion) from tool-tip pose. Curvature and torsion are widely used with motion trajectories for geometric shape matching~\cite{sebastian2003aligning}, in computer vision applications of region-based image segmentation~\cite{schoenemann2011linear}, and surface reconstruction~\cite{strandmark2011curvature}. Few existing works use of curvature and torsion information for gesture and motion segmentation tasks but did so using unsupervised learning~\cite{arn2018motion,despinoy2015unsupervised}. Despoinoy et al.~\cite{despinoy2015unsupervised} showed 94.2\% accuracy in motion segmentation on human movement dataset. Arn et al.~\cite{arn2018motion} used multiple kinematics signals including curvature and torsion to capture surgeons' intentions during a training task. While these works do not explore the effects of kinematics' interaction with vision data, they do show that curvature and torsion, in addition to position and rotation, provide a better understanding of trajectories. 

In this work, we propose combining novel motion invariants with vision and kinematics data to improve gesture recognition. Specifically, we use curvature to measure transitions in the trajectory paths and torsion to measure how much the trajectory path deviates from its osculating plane. We use multi-relational graph network (MRG-Net)~\cite{long2021relational} as our baseline deep learning model to integrate vision, kinematics, and motion invariant measures data for improved gesture recognition. This model offers frame-wise gesture classification to enable real-time gesture recognition without pre-segmented gesture boundaries. We train and validate our model on the JIGSAWS dataset~\cite{gao2014jhu}.
\section{Method} \label{sec:method}

\par We first present the differential geometry approach used to extract the curve of striction from tool-tip trajectories. Second, we describe the computation of our motion invariant measures: curvature and torsion. Third, we present our neural networks model adapted from MRG-Net~\cite{long2021relational} and how we encode the curvature and torsion. The neural network extract visual and kinematics embeddings and process  the joint knowledge of these multi-modal features using a relational graph convolutional network to estimate the gesture prediction on a frame level. 

\subsection{Computing the Striction Curve}

\begin{figure}[htbp]
    \centering
    \includegraphics[width=0.8\columnwidth]{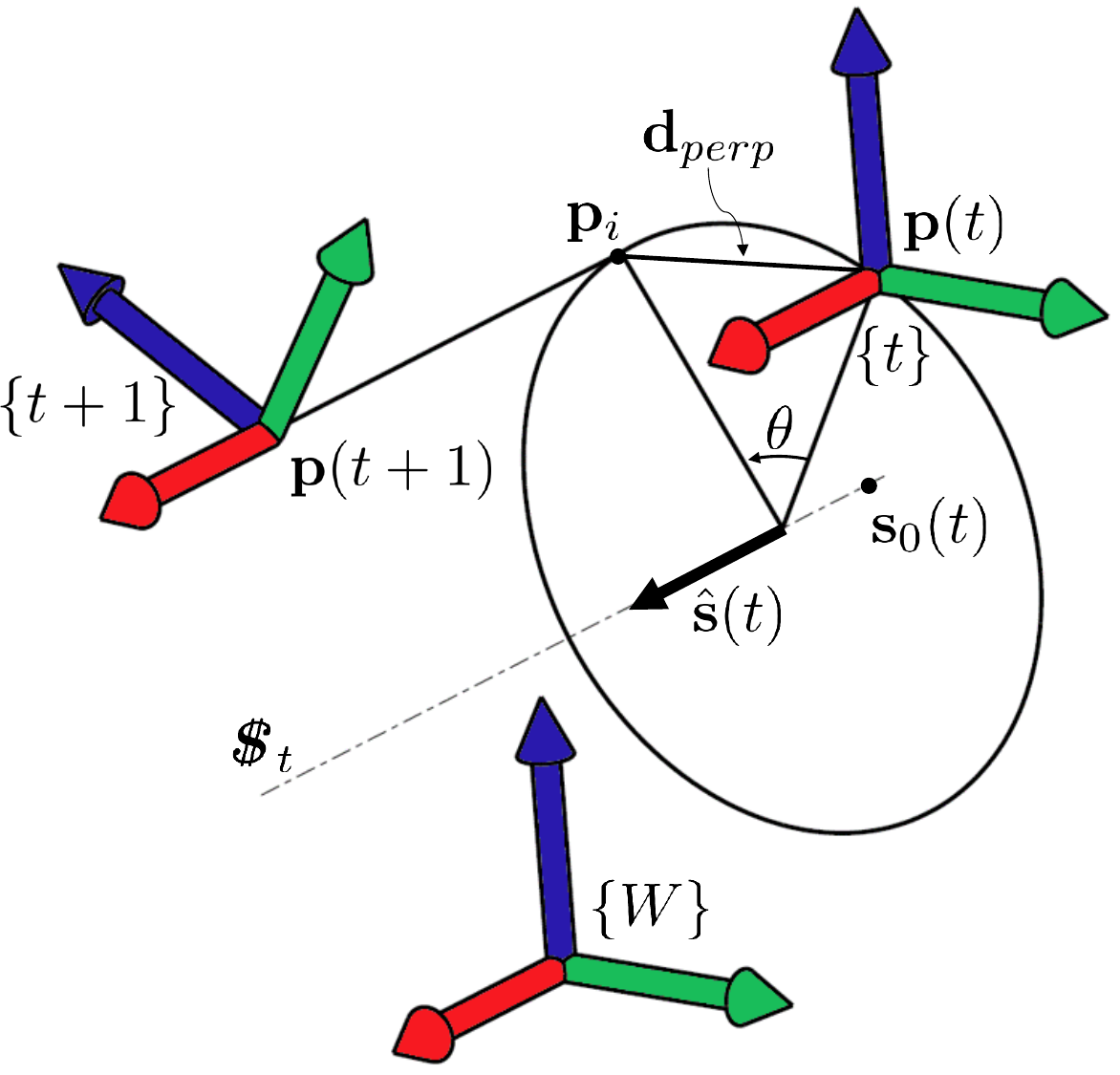}
    \caption{Screw of finite motion $\textit{\textbf{\dollar}}_t$ between two subsequent tool-tip frames. In this figure, $\{W\}$ denotes the world frame, $\{t\}$ denotes the tool-tip frame at time $t$, $\{t+1\}$ denotes the tool-tip frame at time $t+1$, $\mb{s}_0(t)$ denotes the closet point on the screw axis to the world frame origin, and $\uvec{s}(t)$ denotes a unit vector pointing along the direction of the screw axis.}
    \label{fig:screw-terms}
\end{figure}

\begin{figure*}[htbp]
    \centering
    \includegraphics[trim={0cm 7cm 0cm 0cm},clip,width=1\textwidth]{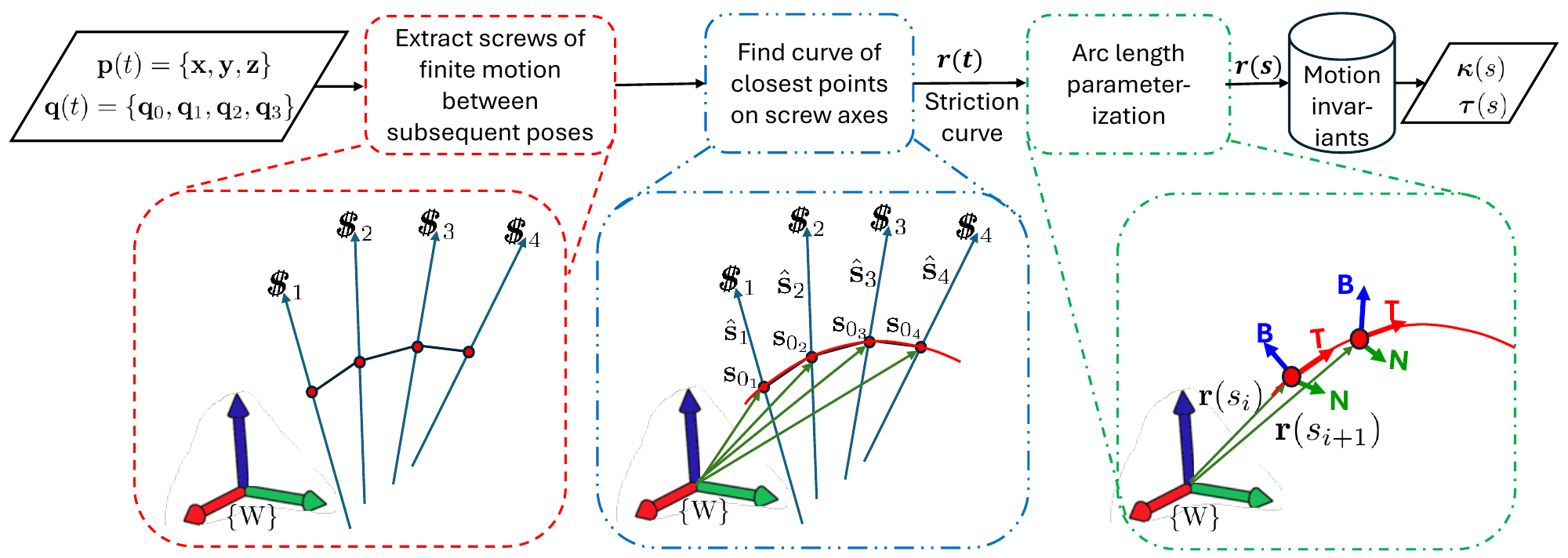}
    \caption{Obtaining motion invariants from tool poses: (a) the screws of finite motion $\textit{\textbf{\dollar}}_i$ between subsequent poses and their common normals, (b) the Pl\"ucker line coordinate parameters $\uvec{s}_i$ and $\mb{s}_{0_i}$ and a spline curve approximating the curve of striction, (c) arc-length parametrization of the striction curve defining the curvature $\kappa(s)$ and torsion $\tau(s)$.}
    \label{fig:kin_pipeline}
\end{figure*}

In this section, we detail how the surgical tool-tip trajectory can be characterized using the \emph{striction curve}. This curve is the line formed from each consecutive closest point along the \emph{screws of finite motion} (SFM) between each successive tool-tip position $\mb{p}\in\realfield{3}$ and rotation $\mb{q}\in\mathbb{H}$ reading \cite{Hunt1990}, shown in Fig.~\ref{fig:screw-terms}. Fig.~\ref{fig:kin_pipeline} shows a conceptual overview of the process of extracting the striction curve from the SFMs.

\par To find this curve, we first extract the SFM between the tool-tip pose at time index $t$ and $t+1$ using Chasle's Theorem \cite{dimentberg1969screw}. As shown in Fig.~\ref{fig:screw-terms}, each SFM can be characterized using 1) the unit vector $\uvec{s}$ pointing along the screw, 2) a point on the screw axis $\mb{s}_0$, 3) the angle $\theta$ rotated about $\uvec{s}$, and 4) a translation along the axis of the screw. 
\par To find $\uvec{s}$, we first find the angular displacement between frames $\{t\}$ and $\{t+1\}$ in the form of a unit quaternion:
\begin{equation}
    \Delta\mb{q}(t) = \mb{q}(t+1) \ostar \mb{q}^{*}(t)
\end{equation}
In this equation, $\mb{q}(t)\in\mathbb{H}$ is the tool rotation at time index $t$, $\mb{q}(t+1)\in\mathbb{H}$ is the tool rotation at time index $t+1$, $\ostar$ represents quaternion multiplication, and $(\cdot)^*$ indicates the conjugate of the quaternion. Now, the axis $\uvec{s}(t)$ can be computed using the normalized complex components of the displacement quaternion:
\begin{equation}
    \mb{s}(t) = \begin{bmatrix} \Delta q_{1}(t) & \Delta q_{2}(t)&
    \Delta q_{3}(t)\end{bmatrix}\T, ~ 
   \uvec{s}(t) = \frac{\mb{s}(t)}{\|\mb{s}(t)\|} 
\end{equation}
Now that we have the direction of the SFM, we must now extract the point on the axis $\mb{s}_0$. To do this, we first compute the angle $\theta$ about the screw axis (see Fig. \ref{fig:screw-terms}) from the real component of the angular displacement quaternion:
\begin{equation}
    \theta(t) = 2\arccos(\Delta q_{0}(t))
\end{equation}
Additionally, we must compute the portion of the frame displacement $\mb{d} = \mb{p}(t+1)-\mb{p}(t)$ that is perpendicular to $\uvec{s}(t)$ which is shown as $\mb{d}_{perp}$ in Fig. \ref{fig:screw-terms}:
\begin{equation}
    \mb{d}_{perp} = \left(\mb{I}- \uvec{s}(t) \uvec{s}(t)\T\right)\mb{d} 
\end{equation}
Now, we can finally compute the point $\mb{s}_0$:
\begin{equation}
    \mb{s}_0 = \frac{1}{2}\left[\mb{p}_i + \mb{p}(t)\right] + \frac{\|\mb{d}_{perp}\|}{2\tan\left(\theta/2\right)}\left[\uvec{s}(t)\times \frac{\mb{d}_{perp}}{\|\mb{d}_{perp}\|}\right]
\end{equation}
Where $\mb{p}_i = \mb{p}(t) + \mb{d}_{perp}$ is the intermediate point shown in Fig. \ref{fig:screw-terms}.  The Pl\"ucker coordinates representing the SFM can now be computed as $\textit{\textbf{\dollar}}_i = [\uvec{s},~\mb{s}_0\times\uvec{s}]\T$ as shown in Fig. \ref{fig:kin_pipeline}.
\par Next, to compute the striction curve, we find the shortest distances connecting two consecutive screw lines by computing the magnitude of the common normal vector $\mb{m}(t)$ joining them. Alg. \ref{alg:shortest_distance}, adopted from~\cite{eikenes2024}, shows how the intersection cases for two lines -- parallel, intersect at a point, or skew -- are handled.

\begin{algorithm}[htbp] 
\caption{Calculate Shortest Distance Between Two Lines}\label{alg:shortest_distance}
\SetKwInput{Input}{Input}
\SetKwInput{Output}{Output}

\Input{Points $\mb{s}_{0}(t), \mb{s}_{0}(t+1)$, direction vectors $\uvec{s}(t), \uvec{s}(t+1)$}
\Output{Closest points $\mb{p}_{a}$, $\mb{p}_{b}$ and shortest distance between the lines}

\BlankLine
\If{$|\|\uvec{s}(t)\|^{2} \|\uvec{s}(t+1)\|^{2} - \uvec{s}(t) \cdot \uvec{s}(t+1)| < \epsilon$}{
    \emph{Lines are parallel:}\\
    Project $\mb{s}_{0}(t)$ onto line defined by $\mb{s}_{0}(t+1)$ and $\uvec{s}(t+1)$:\\
    $A_{\text{proj}} \gets \mb{s}_{0}(t+1) + \left((\mb{s}_{0}(t) - \mb{s}_{0}(t+1)) \uvec{s}\T(t+1)\right) \uvec{s}(t+1)$\;
    $\text{projection} \gets \mb{s}_{0}(t) - A_{\text{proj}}$\;

    \BlankLine
    $\mb{p}_{a} \gets \mb{s}_{0}(t)$\;
    $\mb{p}_{b} \gets A_{\text{proj}}$\;
    $\text{shortestDistance} \gets \|\text{projection}\|$\;
}
\Else{
    \emph{Lines are skew or intersecting:}\\
    $\mu_a \gets \left(\left((\mb{s}_{0}(t+1) - \mb{s}_{0}(t) )\cdot \uvec{s}(t+1) \right)
    \left(\uvec{s}(t) \cdot \uvec{s}(t+1)\right)\right) - 
    \left(\left((\mb{s}_{0}(t+1) - \mb{s}_{0}(t)) \cdot \uvec{s}(t)\right) \|\uvec{s}(t+1)\|^{2}\right) 
    /(\|\uvec{s}(t)\|^{2} \cdot \|\uvec{s}(t+1)\|^{2} - \uvec{s}(t) \cdot \uvec{s}(t+1))$\;
    
    $\mu_b \gets \left(((\mb{s}_{0}(t+1) - \mb{s}_{0}(t)) \cdot \uvec{s}(t+1)\right) + \left(\uvec{s}(t) \cdot \uvec{s}(t+1)\right) \mu_a) / \|\uvec{s}(t+1)\|^{2}$\;
    
    \BlankLine
    $\mb{p}_{a} \gets \mb{s}_{0}(t) + \mu_a  \uvec{s}(t)$\;
    $\mb{p}_{b}\gets \mb{s}_{0}(t+1) + \mu_b  \uvec{s}(t+1) $\;
    $\text{shortestDistance} \gets \|\mb{p}_{b} - \mb{p}_{a}\|$\;
}
\Return{$\text{shortestDistance}, \mb{p}_{a}, \mb{p}_{b}$}
\end{algorithm}
The algorithm takes two points and two direction vectors (e.g., $\mb{s}_{0}(t)$ and $\uvec{s}(t)$) and returns the shortest distance by computing the magnitude of the perpendicular vector $\|\mb{u}(t)\|$ between the lines and return the intersection points. 

\par Next, we compute the summation of all distances $\sum_{n \in N-1}\|\mb{u_{n}}(t)\|$, which provides us with the total arc length needed to parametrize the striction curve. We define the intersection points between $\mb{u}(t)$ and screw lines with $\mb{j}$ where $\mb{j} \in \realfield{3}$ and $\mb{j}_{n}$ represents the intersection point with the nth screw line where $n =\{1, 2, ..N\}$.

\par In order to parameterize the striction curve by its arc length, we interpolate between the intersection points by passing a cubic spline. We then resampled the spline curve at equal distances given by $\frac{\sum_{n \in N-1}\|\mb{u_{n}}(t)\|}{\text{N-1}}$.

\subsection{Curvature and Torsion of the Striction Curve}
%
In the previous section, we discussed how the tool-tip's motion can be characterized using the striction curve. Although users may exhibit different striction curves, we hypothesize that the surgical gestures can be identified in part by using special parameters of the curve called the \emph{curvature} and \emph{torsion}. In this section, we will define these parameters and detail how they are calculated for the robotic tool-tip striction curves. 

 %
\par For a differentiable curve $\mb{r}(s)\in\realfield{3}$ that has been parameterized by its arc length $s\in\realfield{+}$, we can assign three unit vectors $\uvec{t}(s) = \mb{r}'(s)/\|\mb{r}'(s)\|$ (the tangent vector), $\uvec{n}(s) = \mb{r}''(s)/\|\mb{r}''(s)\|$ (the normal vector), and $\uvec{b}(s) = \uvec{t}(s) \times \uvec{n}(s)$ (the bi-normal vector) that define an orthonormal basis ${}^w\mb{R}(s) = [\uvec{n}(s),~\uvec{b}(s),~\uvec{t}(s)] \in SO(3)$ at every point along the curve. These unit vectors evolve with respect to arc-length using the Frenet-Serret equations \cite{nutbourne1988differential}:  
 \begin{equation}\label{eq:frenet-serret}
\begin{aligned}
    \uvec{t}'(s) &= \kappa(s)\uvec{n}(s) \\
    \uvec{n}'(s) &= -\kappa(s)\uvec{t}(s)  + \tau(s)\uvec{b}(s)\\
    \uvec{b}'(s) &= -\tau(s)\uvec{n}(s)
\end{aligned}
\end{equation}
 In these equations, $\kappa(s)$ and $\tau(s)$ are the curvature and torsion of $\mb{r}(s)$, respectively. The curvature can be interpreted as the the magnitude of the rate of change of the tangent vector and the torsion measures the rate of change of the bi-normal vector. For a straight curve, the curvature will be zero. For a planar curve, the torsion will be zero.
 \par As shown in Alg. \ref{alg:signed_curvature}, the magnitude of the curvature is equal to the magnitude of $\mb{r}''(s)$. To compute the sign of the curvature, we define $\uvec{n}$ such that it starts in the same direction as $\mb{r}''(s)$. However, at inflection points, $\mb{r}''(s)$ will suddenly flip to the opposite side of the curve. To detect this change, we test if the dot product between the current and previous values of $\mb{r}''(s)$ is negative. At these points, we change the sign of the curvature (denoted $\gamma$ in Alg. \ref{alg:signed_curvature}) such that $\uvec{n}$ does not also flip sides \cite{nutbourne1988differential}. We maintain the value of $\gamma$ until another inflection point has been reached.    

%
\begin{algorithm}
\caption{Signed Curvature}\label{alg:signed_curvature}
\small
$s \gets ds$\;
$\gamma \gets 1$;\Comment{Sign of $\kappa$}\\ 
$\kappa(0) \gets \|\mb{r}''(0)\|$\;
\While{$s \leq s_{max}$}
{
    \If(\Comment{$\mb{r}''(s)$ has flipped}){$\mb{r}''(s)\T\mb{r}''(s-ds) < 0$}  
     { 
        $\gamma \gets -\gamma$;
     }
    $\kappa(s) \gets \gamma\|\mb{r}''(s)\|$\;
    $s \gets s+ds$;
}
\Return{$\kappa(s),  s = [0, s_{max}]$}
\end{algorithm}
\par The torsion can be computed from the determinate of a matrix whose columns are the first, second, and third derivatives of $\mb{r}(s)$ with respect the the arc length $s$ and the curvature.
\begin{equation}\label{eq:torsion}
    \tau(s) = \frac{1}{\kappa^2(s)}\det\left[\mb{r}'(s),~\mb{r}''(s),~ \mb{r}'''(s)\right]
\end{equation}
Because the curvature is zero for straight curves, \eqref{eq:torsion} is undefined for straight portions of the striction curve. To deal with these numerical issues, we define the torsion to be zero when the curvature is very small. 

\subsection{Graph Neural Network for Gesture Recognition}
\par In the previous section, we discussed how the curvature and torsion can be extracted from the striction curve characterizing the tool-tip's trajectory. In this section, we present the machine learning approach for extracting kinematic and motion invariant measures features and using them with vision data to recognize surgical gestures. 

\par We adopt the same kinematic feature extractor as in~\cite{long2021relational}, with two parallel feature extractors consisting of a temporal convolutional network (TCN) and a long-short-term memory (LSTM). The TCN extracts the short-term temporal dependencies and LSTM extracts the long-term temporal dependencies. The long and short temporal dependencies are averaged to obtain the final temporal kinematic features vector. This feature encoding is done for each robotic arm since each arm can be used to perform different parts of the surgical task. For visual features, we replace the ResNet features in MRG-Net with features from~\cite{dipietro2016recognizing} since they were shown to perform better in both~\cite{weerasinghe2024multimodal} and our own preliminary experiments.

\begin{figure}[htbp]
    \centering
    \includegraphics[trim={0cm 0cm 0cm 0cm},clip,width=\columnwidth]{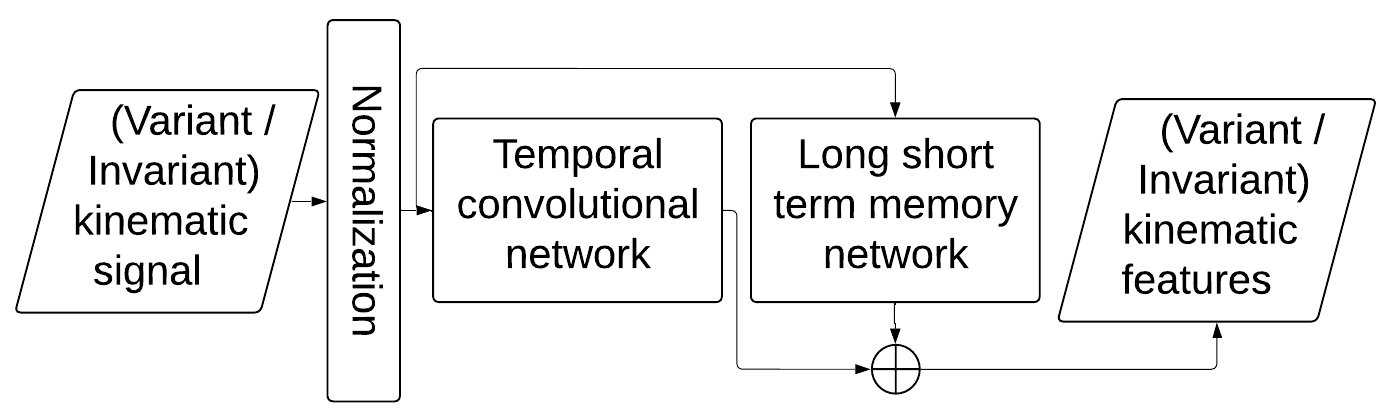}
    \caption{Network architecture used to extract the temporal feature from the kinematics signal. Variant kinematics refers to the tool-tip poses while invariant kinematics refers to the computed curvature and torsion. The outputs of the TCN and LSTM network are averaged.}
    \label{fig:pipeline_part1}
\end{figure}

\par Next, we combine the feature vectors from the vision and kinematics. We use the graph learning module proposed in~\cite{long2021relational} to fuse the features at each time step. Graph learning provides a structured interaction between features from each modality to provide accurate gesture recognition. 
\begin{figure}[htbp]
    \centering
    \includegraphics[width=\columnwidth]{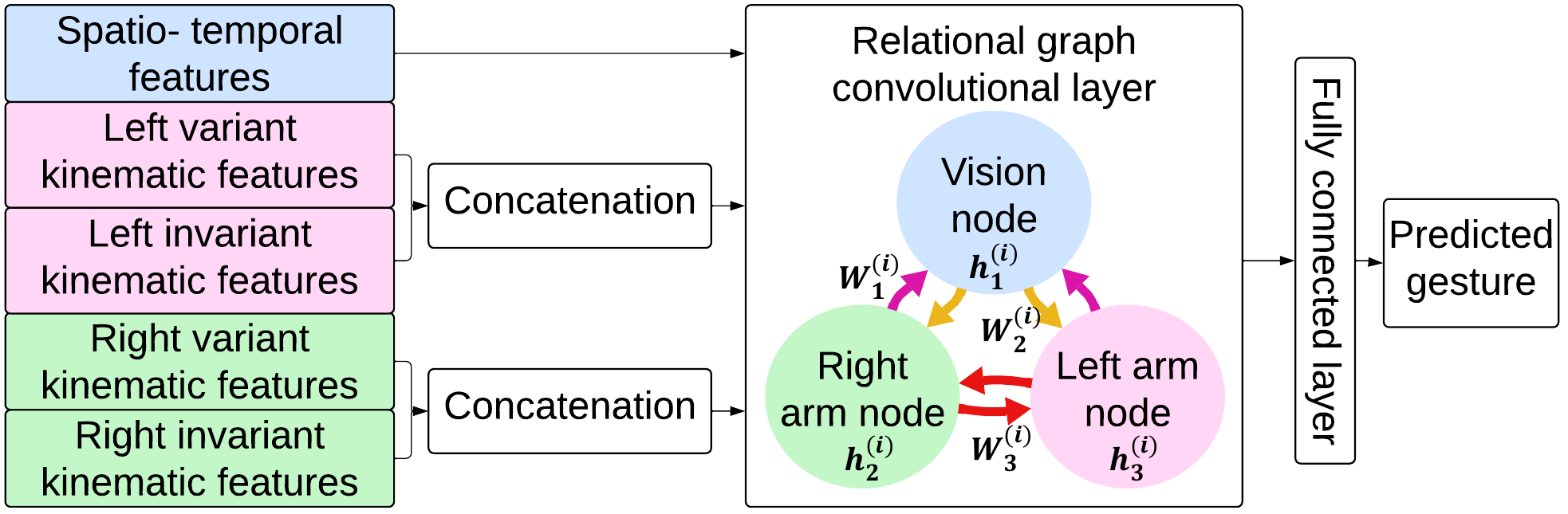}
    \caption{Overall relational graph network showing three nodes for vision, and left and right kinematics. We show a single graph convolutional layer to show the interactions between nodes, adapted from~\cite{long2021relational}. The variant and invariant features (output from the network in Fig.~\ref{fig:pipeline_part1}) are concatenated and passed to the corresponding arm node. The arrows represent the relations between different nodes. After message passing, the hidden features are passed to a fully connected network to estimate the gesture per frame.}
    \label{fig:pipeline_part2}
\end{figure}

\par Fig. \ref{fig:pipeline_part2} shows how the visual and kinematic features are constructed into a graph representation.  Each graph node is updated through message aggregation from neighboring nodes following an aggregation rule given in:
\begin{equation} \label{eq:mesage_passing}
    h_{i}^{(l+1)} = \sigma(\sum_{j \in \mathcal{N}_{i}} f_{m}(h_{i}^{l},h_{j}^{l})), 
\end{equation}
where $h_{i}$ is the hidden state of the node $v_{i}$ in the $l$-th graph network layer, $\mathcal{N}_{i}$ is the set of all indices connected to node $i$, $f_{m}(\cdot,\cdot)$ is the message accumulation function, and $\sigma(\cdot)$ is the element-wise non-linear activation function. 
This message passage network has been shown to impose a learnable message-passing process for the nodes with shared information. Robotic surgery contains plenty of complementary interactions between ``what you see'' and ``what you do''. Effectively identifying the inherent useful relationships among them is difficult while critical to boosting the performance of gesture recognition.

\subsubsection*{Loss Function}

Gesture classes are imbalanced in the JIGSAWS dataset, particularly on a frame level. Different gestures appear with different frequency and duration. Thus we use the weighted cross-entropy to address the imbalance within training samples. With $\alpha$ as the class balancing weight, $\Theta$ as MRG-Net parameters of all trainable layers, and $\hat{c}_{t}$ as the classification prediction of each frame, we optimize the overall loss function:
\begin{equation} \label{eq:Weighted_CEloss}
 \mathcal{L}(\mathcal{X}; \mathcal{Y}; \Theta) = \frac{1}{T} \sum_{t} - \alpha \log \hat{c}_{t},
\end{equation}
where $\mathcal{X}$ is the multi-modal input space, and $\mathcal{Y}$ denotes the gesture categories.

\section{Experimental Setup}\label{sec:experiment}

\subsection{Network Architecture and Hyperparameters}

First, to extract the motion invariant measures for each arm, we pass the position and quaternion into the pipeline shown in Fig. \ref{fig:kin_pipeline}. The tool pose and shape invariant measures are normalized to zero mean and unit variance, and then passed into the kinematic feature extractor, as shown in Fig.~\ref{fig:pipeline_part1}. The TCN block consists of 3 temporal convolutional layers and encoder and decoder backbone with $\{64, 96, 128\}$ filters for encoder and $\{96, 64, 64\}$ filters for decoder with a kernel size of 51. The LSTM has a single hidden layer and a hidden size of 128. The resulting features from LSTM and TCN are averaged and passed to the next stage of the network.

The proposed framework uses relational graph layers, implemented with the Deep Graph Library (DGL)~\cite{wang2019deep}, to share information among kinematic and spatio-temporal features from endoscopic video. The relational graph layers employ 64-dimensional hidden and output states with a 0.2 dropout rate and no hidden layers. We used the same hyper-parameters reported in~\cite{long2021relational}. We used the Adam~\cite{diederik2014adam} optimizer with a learning rate of $1e^{\text{-}3}$ with a learning rate scheduler and weight decay of $1e^{\text{-}4}$ to train the proposed network. We ablate over different combinations of kinematic representations to show the effect of tool-tip pose and motion invariant measures on gesture recognition. 
\subsection{Metrics and Dataset}
\par We evaluated using the publicly available JIGSAWS dataset~\cite{gao2014jhu} and the results are reported on the suturing dataset. All experiments were conducted with an Intel Core i9-13900K, 64GB DRAM, and an NVIDIA GeForce RTX 4090 running Ubuntu 20.04.6 LTS. Each experiment trained for 16 hours over 150 steps, with validation at each step.
\par The kinematic data for each video includes information about the tool-tip position and rotation. Each frame of the suturing is labeled with one of ten gestures (see~\cite{gao2014jhu} for details). We used a leave-one-user-out cross-validation. To evaluate our model's performance, we calculated the percentage of correctly classified frames (accuracy) and the Edit Score~\cite{lea2016segmental} (which has a range of [0, 100], the higher the score the better).
\section{Evaluation}  \label{sec:evlauation}

\begin{figure*}[htb]
    \centering
    \includegraphics[trim={0cm 0cm 0cm 0cm},clip,width=\textwidth]{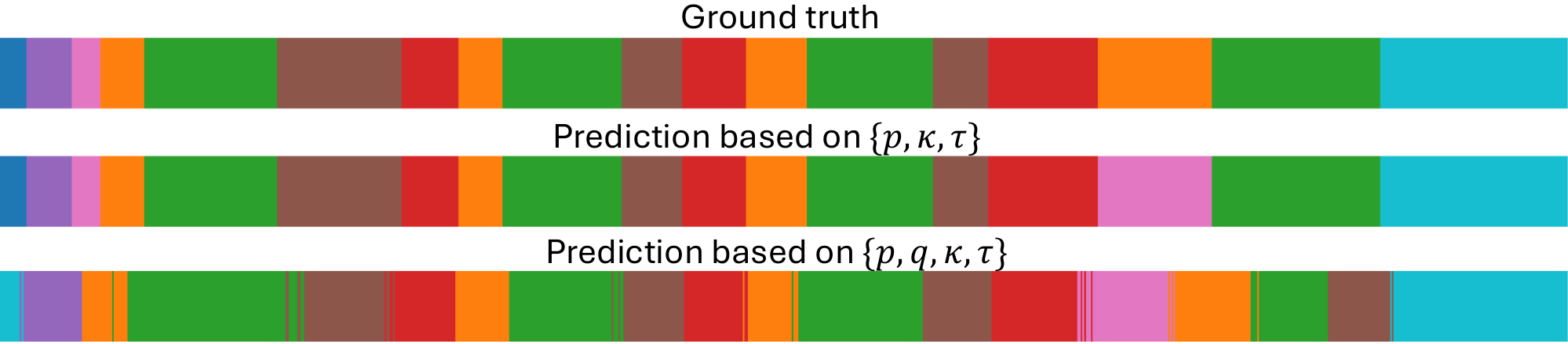}
    \caption{The color-coded ribbons compare the two ablations with the highest and lowest gesture recognition accuracy: the highest from tool-tip position, curvature, and torsion, and the lowest from tool-tip position, rotation, curvature, and torsion.}
    \label{fig:qualitative_trial}
\end{figure*}

\par In Table \ref{table:concat_ablation}, we show the results of the proposed gesture recognition pipeline. We also show the ablation study over the impact of using different kinematic representations. We observe that concatenating the position signal along with curvature and torsion obtains the highest gesture recognition accuracy ($90.3 \pm 5.1$) and edit score ($89.0 \pm 5.5$). This suggests that curvature and torsion encode additional information compared to the pose data, supporting our hypothesis that motion invariant measures can aid surgical gesture recognition. 

\begin{table}[h!]
\centering 
\caption{Performance of gesture recognition in the ablation study of different kinematic signals for the suturing task. A set represents the feature concatenated together for the kinematic node for each arm. Note the \{p, q\} set is the same input as the MRG-Net and we obtain similar results as reported in~\cite{long2021relational}}
\label{table:concat_ablation}
\begin{tabularx}{\columnwidth}{>{\centering\arraybackslash}X >{\centering\arraybackslash}X >{\centering\arraybackslash}X}
\hline
\textbf{Kinematic features} & \textbf{Accuracy (std. dev.) (\%)} & \textbf{Edit Score (std. dev.)  (\%)} \\ \hline
\{p\} & 88.8 $\pm$ 5.5 & 81.5 $\pm$ 19.5 \\
\{p, q\} & 87.0 $\pm$ 6.3 & 85.6 $\pm$ 5.7 \\
\{p,  q, $\kappa$, $\tau$\}  & 76.3 $\pm$ 5.7 & 29.0 $\pm$ 4.5 \\
\{p, $\kappa$, $\tau$\} & \textbf{90.3} $\pm$ \textbf{5.1} & \textbf{89.0} $\pm$ \textbf{5.5}  \\ \hline
\multicolumn{3}{l}{
p and q represent position and quaternion features respectively.}\\
\multicolumn{3}{l}{$\kappa$ and $\tau$ represent curvature and torsion features respectively.}
\end{tabularx}  
\end{table}

Interestingly, the accuracy decreases when quaternion is included. A recent work reported that quaternions are a difficult representation for neural networks to learn over~\cite{geist2024learning}, which may contribute to the decrease in accuracy observed here. Furthermore, using quaternion, curvature, and torsion together resulted in a significant decrease compared to all other ablations. We speculate that this could be due to repetitive information between the rotation, curvature, and torsion reducing the efficiency in learning. 
Fig.~\ref{fig:qualitative_trial} show qualitative results for gesture recognition for the best and worst ablations, $\{p, \kappa, \tau\}$ and $\{p, q, \kappa, \tau\}$. We observe that when the quaternion is not included, the gestures are often recognized correctly and the transitions between gestures are detected at the right time. Most errors derive from a single misclassified gesture. However, the network's ability to detect transitions and gestures correctly reduces significantly when the quaternion is included.

\begin{figure*}[htb]
    \centering
    \includegraphics[trim={0cm 0cm 0cm 0cm},clip,width=\textwidth]{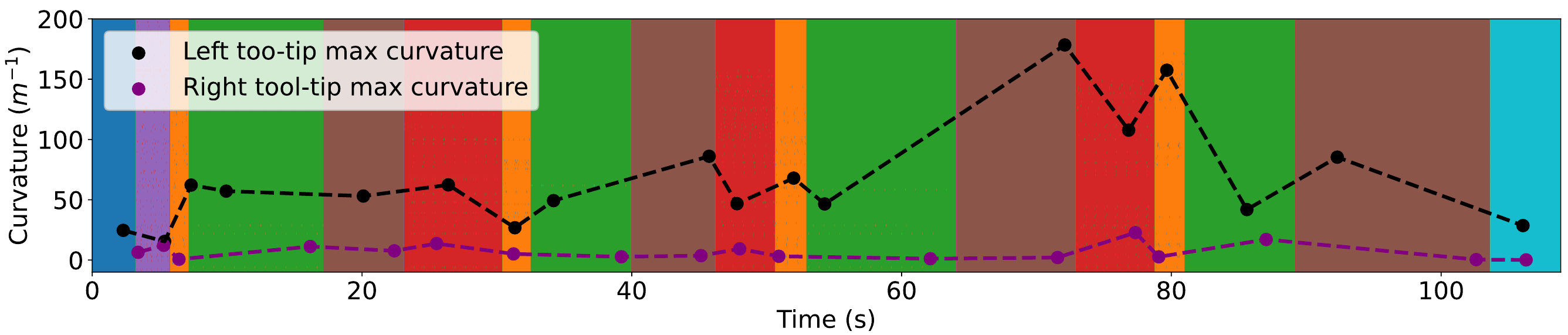}
    \caption{The maximum curvatures resulting from left and right tool-tip trajectories are plotted per gesture. The color-coded ribbon illustrates that curvature inflection points happen at gesture boundaries.}
    \label{fig:curvature}
\end{figure*}

To further explore the contribution of curvature to gesture recognition and to develop an intuitive understanding of why it may improve results, we plot the curvature against gestures for one suturing trial in Fig.~\ref{fig:curvature}. We observe that inflection points in curvature, either for the right or left tool, often align with gesture boundaries. This may make intuitive sense as direction changes of tools often indicate a new gesture. 

\section{Conclusions and Future Work}

In this paper, we introduce novel motion invariant features to improve gesture recognition for surgical tasks. We use a multi-modal graph neural network, MRG-Net, to model gestures by first encoding vision and the kinematic features associated with each arm, and then sharing information between each data source through graph message passing. We ablate over different combinations of the kinematic features: position, rotation, curvature, and torsion. We test the model's performance on the JIGSAWS suturing dataset. Our motion invariant features improve upon the state-of-art gesture recognition performance achieving an accuracy of 90.3\%. We show that local rotation information provided by curvature and torsion signals outperforms the traditional quaternion representation in gesture recognition tasks. This supports prior works' claims on the limitation of the current rotation representations~\cite{geist2024learning} and provides a new way to represent rotation signals and embed geometric information into the learned representation.

\par Though our investigation has shown improvement in gesture recognition when adding curvature and torsion signals, there are still some limitations stemming from approximations used in this work. The traditional notion of the axode of motion uses the instantaneous screws of motion. In this work, we approximated the axode of motion by using the screws of finite displacement between two subsequent poses (position \& rotation) of the tools. The underlying assumption behind this approximation is that the sampling rate of the tool position measurements is relatively fast. The motivation for this approximation stems from the need to avoid computing screws of instantaneous motion by using linear and angular velocities from finite-difference pose information since such calculations become very sensitive to measurement noise as the sampling period becomes small. We observe that the neural network is more prone to overfitting with quaternions, a challenge that becomes even more pronounced when additional features are introduced. This is reflected in the edit score for the ${p, q, \kappa, \tau}$ trial, shown in Table~\ref{table:concat_ablation}. We observe from Fig.~\ref{fig:qualitative_trial} that the trial with quaternions estimated frequent transitions, which disrupted gesture sequencing. This led to more sequence mismatches, reducing the edit score significantly. JIGSAWS's limited dataset size may contribute to overfitting. To further demonstrate the generalizability of motion invariants for gesture recognition, future work should evaluate this method on other datasets (e.g., PETRAW~\cite{huaulme2022peg}). 

\par While tool-tip velocity and gripper angle can additionally aid gesture recognition, this work focuses on the effects of motion invariant features on neural-network based gesture representation. Additional information about tool-tip velocity and gripper information may improve gesture recognition results. We focus on using motion invariants to develop on gesture-specific skill assessment. Motion invariants represent how surgeons move their instruments, which may be highly indicative of skill. Gesture-specific skill assessment would help provide feedback on which gestures a particular user finds challenging, and create opportunity to provide targeted practice. In addition to refining our kinematic features, we plan to explore different network architectures to combine the information from the motion invariants. Recent works have indicated that intermediate fusion may be more beneficial for gesture recognition than late fusion~\cite{boulahia2021early}. 

By using motion invariant features, we achieved state-of-the-art results in gesture recognition on the JIGSAWS dataset. We show that the novel curvature and torsion features offer complementary information to the vision and raw kinematics data. Our results suggest that the curvature and torsion may improve efficiency of training and reduce overfitting. As our gesture-recognition pipeline is frame-wise and does not require pre-segmented gestures, it can be integrated into downstream tasks such as immediate skill assessment and feedback, and automation.




\bibliographystyle{IEEEtran}
\bibliography{bib/references}

\begin{thebibliography}{10}
\providecommand{\url}[1]{#1}
\csname url@rmstyle\endcsname
\providecommand{\newblock}{\relax}
\providecommand{\bibinfo}[2]{#2}
\providecommand\BIBentrySTDinterwordspacing{\spaceskip=0pt\relax}
\providecommand\BIBentryALTinterwordstretchfactor{4}
\providecommand\BIBentryALTinterwordspacing{\spaceskip=\fontdimen2\font plus
\BIBentryALTinterwordstretchfactor\fontdimen3\font minus \fontdimen4\font\relax}
\providecommand\BIBforeignlanguage[2]{{%
\expandafter\ifx\csname l@#1\endcsname\relax
\typeout{** WARNING: IEEEtran.bst: No hyphenation pattern has been}%
\typeout{** loaded for the language `#1'. Using the pattern for}%
\typeout{** the default language instead.}%
\else
\language=\csname l@#1\endcsname
\fi
#2}}

\bibitem{probst2023review}
P.~Probst, ``A review of the role of robotics in surgery: To davinci and beyond!'' \emph{Missouri medicine}, vol. 120, no.~5, p. 389, 2023.

\bibitem{guthart2000intuitive}
G.~S. Guthart and J.~K. Salisbury, ``The intuitive/sup tm/telesurgery system: overview and application,'' in \emph{Proceedings 2000 ICRA. Millennium Conference. IEEE International Conference on Robotics and Automation. Symposia Proceedings (Cat. No. 00CH37065)}, vol.~1.\hskip 1em plus 0.5em minus 0.4em\relax IEEE, 2000, pp. 618--621.

\bibitem{tao2012sparse}
L.~Tao, E.~Elhamifar, S.~Khudanpur, G.~D. Hager, and R.~Vidal, ``Sparse hidden markov models for surgical gesture classification and skill evaluation,'' in \emph{Information Processing in Computer-Assisted Interventions: Third International Conference, IPCAI 2012, Pisa, Italy, June 27, 2012. Proceedings 3}.\hskip 1em plus 0.5em minus 0.4em\relax Springer, 2012, pp. 167--177.

\bibitem{long2021relational}
Y.~Long, J.~Y. Wu, B.~Lu, Y.~Jin, M.~Unberath, Y.-H. Liu, P.~A. Heng, and Q.~Dou, ``Relational graph learning on visual and kinematics embeddings for accurate gesture recognition in robotic surgery,'' in \emph{2021 IEEE International Conference on Robotics and Automation (ICRA)}.\hskip 1em plus 0.5em minus 0.4em\relax IEEE, 2021, pp. 13\,346--13\,353.

\bibitem{lea2016temporal}
C.~Lea, R.~Vidal, A.~Reiter, and G.~D. Hager, ``Temporal convolutional networks: A unified approach to action segmentation,'' in \emph{Computer Vision--ECCV 2016 Workshops: Amsterdam, The Netherlands, October 8-10 and 15-16, 2016, Proceedings, Part III 14}.\hskip 1em plus 0.5em minus 0.4em\relax Springer, 2016, pp. 47--54.

\bibitem{wu2021cross}
J.~Y. Wu, A.~Tamhane, P.~Kazanzides, and M.~Unberath, ``Cross-modal self-supervised representation learning for gesture and skill recognition in robotic surgery,'' \emph{International Journal of Computer Assisted Radiology and Surgery}, vol.~16, pp. 779--787, 2021.

\bibitem{van2020multi}
B.~van Amsterdam, M.~J. Clarkson, and D.~Stoyanov, ``Multi-task recurrent neural network for surgical gesture recognition and progress prediction,'' in \emph{2020 IEEE international conference on robotics and automation (ICRA)}.\hskip 1em plus 0.5em minus 0.4em\relax IEEE, 2020, pp. 1380--1386.

\bibitem{kiyasseh2023vision}
D.~Kiyasseh, R.~Ma, T.~F. Haque, B.~J. Miles, C.~Wagner, D.~A. Donoho, A.~Anandkumar, and A.~J. Hung, ``A vision transformer for decoding surgeon activity from surgical videos,'' \emph{Nature biomedical engineering}, vol.~7, no.~6, pp. 780--796, 2023.

\bibitem{van2021gesture}
B.~van Amsterdam, M.~J. Clarkson, and D.~Stoyanov, ``Gesture recognition in robotic surgery: a review,'' \emph{IEEE Transactions on Biomedical Engineering}, vol.~68, no.~6, 2021.

\bibitem{qin2020temporal}
Y.~Qin, S.~A. Pedram, S.~Feyzabadi, M.~Allan, A.~J. McLeod, J.~W. Burdick, and M.~Azizian, ``Temporal segmentation of surgical sub-tasks through deep learning with multiple data sources,'' in \emph{2020 IEEE International Conference on Robotics and Automation (ICRA)}.\hskip 1em plus 0.5em minus 0.4em\relax IEEE, 2020, pp. 371--377.

\bibitem{hutchinson2023towards}
K.~Hutchinson, Z.~Li, I.~Reyes, and H.~Alemzadeh, ``Towards surgical context inference and translation to gestures,'' in \emph{2023 IEEE International Conference on Robotics and Automation (ICRA)}.\hskip 1em plus 0.5em minus 0.4em\relax IEEE, 2023, pp. 6802--6809.

\bibitem{geist2024learning}
A.~R. Geist, J.~Frey, M.~Zobro, A.~Levina, and G.~Martius, ``Learning with 3d rotations, a hitchhiker's guide to so (3),'' \emph{arXiv preprint arXiv:2404.11735}, 2024.

\bibitem{gao2014jhu}
Y.~Gao, S.~S. Vedula, C.~E. Reiley, N.~Ahmidi, B.~Varadarajan, H.~C. Lin, L.~Tao, L.~Zappella, B.~B{\'e}jar, D.~D. Yuh, \emph{et~al.}, ``Jhu-isi gesture and skill assessment working set (jigsaws): A surgical activity dataset for human motion modeling,'' in \emph{MICCAI workshop: M2cai}, vol.~3, no. 2014, 2014, p.~3.

\bibitem{weerasinghe2024multimodal}
K.~Weerasinghe, S.~H.~R. Roodabeh, K.~Hutchinson, and H.~Alemzadeh, ``Multimodal transformers for real-time surgical activity prediction,'' \emph{arXiv preprint arXiv:2403.06705}, 2024.

\bibitem{chen2005formulas}
C.-H. Chen, ``Formulas for computing axode normal and curvatures with derivation based on fundamental concepts of engineering differential geometry,'' \emph{Mechanism and machine theory}, vol.~40, no.~7, pp. 834--848, 2005.

\bibitem{boutin2000numerically}
M.~Boutin, ``Numerically invariant signature curves,'' \emph{International Journal of Computer Vision}, vol.~40, pp. 235--248, 2000.

\bibitem{calabi1998differential}
E.~Calabi, P.~J. Olver, C.~Shakiban, A.~Tannenbaum, and S.~Haker, ``Differential and numerically invariant signature curves applied to object recognition,'' \emph{International Journal of Computer Vision}, vol.~26, pp. 107--135, 1998.

\bibitem{liu2021towards}
D.~Liu, Q.~Li, T.~Jiang, Y.~Wang, R.~Miao, F.~Shan, and Z.~Li, ``Towards unified surgical skill assessment,'' in \emph{Proceedings of the IEEE/CVF conference on computer vision and pattern recognition}, 2021, pp. 9522--9531.

\bibitem{zappella2013surgical}
L.~Zappella, B.~B{\'e}jar, G.~Hager, and R.~Vidal, ``Surgical gesture classification from video and kinematic data,'' \emph{Medical image analysis}, vol.~17, no.~7, pp. 732--745, 2013.

\bibitem{gurcan2019surgical}
I.~Gurcan and H.~Van~Nguyen, ``Surgical activities recognition using multi-scale recurrent networks,'' in \emph{ICASSP 2019-2019 IEEE International Conference on Acoustics, Speech and Signal Processing (ICASSP)}.\hskip 1em plus 0.5em minus 0.4em\relax IEEE, 2019, pp. 2887--2891.

\bibitem{shi2022recognition}
C.~Shi, Y.~Zheng, and A.~M. Fey, ``Recognition and prediction of surgical gestures and trajectories using transformer models in robot-assisted surgery,'' in \emph{2022 IEEE/RSJ International Conference on Intelligent Robots and Systems (IROS)}.\hskip 1em plus 0.5em minus 0.4em\relax IEEE, 2022, pp. 8017--8024.

\bibitem{li2023robotic}
Z.~Li, I.~Reyes, and H.~Alemzadeh, ``Robotic scene segmentation with memory network for runtime surgical context inference,'' in \emph{2023 IEEE/RSJ International Conference on Intelligent Robots and Systems (IROS)}.\hskip 1em plus 0.5em minus 0.4em\relax IEEE, 2023, pp. 6601--6607.

\bibitem{agarwal2022temporal}
S.~Agarwal, C.~S. Pradeep, and N.~Sinha, ``Temporal surgical gesture segmentation and classification in multi-gesture robotic surgery using fine-tuned features and calibrated ms-tcn,'' in \emph{2022 IEEE International Conference on Signal Processing and Communications (SPCOM)}.\hskip 1em plus 0.5em minus 0.4em\relax IEEE, 2022, pp. 1--5.

\bibitem{sebastian2003aligning}
T.~B. Sebastian, P.~N. Klein, and B.~B. Kimia, ``On aligning curves,'' \emph{IEEE transactions on pattern analysis and machine intelligence}, vol.~25, no.~1, pp. 116--125, 2003.

\bibitem{schoenemann2011linear}
T.~Schoenemann, F.~Kahl, S.~Masnou, and D.~Cremers, ``A linear framework for region-based image segmentation and inpainting involving curvature penalization,'' \emph{arXiv preprint arXiv:1102.3830}, 2011.

\bibitem{strandmark2011curvature}
P.~Strandmark and F.~Kahl, ``Curvature regularization for curves and surfaces in a global optimization framework,'' in \emph{International workshop on energy minimization methods in computer vision and pattern recognition}.\hskip 1em plus 0.5em minus 0.4em\relax Springer, 2011, pp. 205--218.

\bibitem{arn2018motion}
R.~T. Arn, P.~Narayana, T.~Emerson, B.~A. Draper, M.~Kirby, and C.~Peterson, ``Motion segmentation via generalized curvatures,'' \emph{IEEE transactions on pattern analysis and machine intelligence}, vol.~41, no.~12, pp. 2919--2932, 2018.

\bibitem{despinoy2015unsupervised}
F.~Despinoy, D.~Bouget, G.~Forestier, C.~Penet, N.~Zemiti, P.~Poignet, and P.~Jannin, ``Unsupervised trajectory segmentation for surgical gesture recognition in robotic training,'' \emph{IEEE Transactions on Biomedical Engineering}, vol.~63, no.~6, pp. 1280--1291, 2015.

\bibitem{Hunt1990}
K.~H. Hunt, \emph{\BIBforeignlanguage{en}{Kinematic geometry of mechanisms}}, ser. Engineering Science S.\hskip 1em plus 0.5em minus 0.4em\relax Oxford, England: Clarendon Press, Apr. 1990.

\bibitem{dimentberg1969screw}
F.~Dimentberg, \emph{The screw calculus and its applications in mechanics}.\hskip 1em plus 0.5em minus 0.4em\relax Foreign Technology Division, 1969.

\bibitem{eikenes2024}
\BIBentryALTinterwordspacing
A.~Eikenes. (2024) Intersection point of lines in 3d space. MATLAB Central File Exchange. Retrieved November 15, 2024. [Online]. Available: \url{https://www.mathworks.com/matlabcentral/fileexchange/37192-intersection-point-of-lines-in-3d-space}
\BIBentrySTDinterwordspacing

\bibitem{nutbourne1988differential}
A.~W. Nutbourne and R.~R. Martin, \emph{Differential Geometry Applied to Curve and Surface Design}.\hskip 1em plus 0.5em minus 0.4em\relax Ellis Horwood, Ltd., 1988.

\bibitem{dipietro2016recognizing}
R.~DiPietro, C.~Lea, A.~Malpani, N.~Ahmidi, S.~S. Vedula, G.~I. Lee, M.~R. Lee, and G.~D. Hager, ``Recognizing surgical activities with recurrent neural networks,'' in \emph{Medical Image Computing and Computer-Assisted Intervention--MICCAI 2016: 19th International Conference, Athens, Greece, October 17-21, 2016, Proceedings, Part I 19}.\hskip 1em plus 0.5em minus 0.4em\relax Springer, 2016, pp. 551--558.

\bibitem{wang2019deep}
M.~Y. Wang, ``Deep graph library: Towards efficient and scalable deep learning on graphs,'' in \emph{ICLR workshop on representation learning on graphs and manifolds}, 2019.

\bibitem{diederik2014adam}
P.~K. Diederik, ``Adam: A method for stochastic optimization,'' \emph{(No Title)}, 2014.

\bibitem{lea2016segmental}
C.~Lea, A.~Reiter, R.~Vidal, and G.~D. Hager, ``Segmental spatiotemporal cnns for fine-grained action segmentation,'' in \emph{Computer Vision--ECCV 2016: 14th European Conference, Amsterdam, The Netherlands, October 11-14, 2016, Proceedings, Part III 14}.\hskip 1em plus 0.5em minus 0.4em\relax Springer, 2016, pp. 36--52.

\bibitem{huaulme2022peg}
A.~Huaulm{\'e}, K.~Harada, Q.-M. Nguyen, B.~Park, S.~Hong, M.-K. Choi, M.~Peven, Y.~Li, Y.~Long, Q.~Dou, \emph{et~al.}, ``Peg transfer workflow recognition challenge report: Does multi-modal data improve recognition?'' \emph{arXiv preprint arXiv:2202.05821}, 2022.

\bibitem{boulahia2021early}
S.~Y. Boulahia, A.~Amamra, M.~R. Madi, and S.~Daikh, ``Early, intermediate and late fusion strategies for robust deep learning-based multimodal action recognition,'' \emph{Machine Vision and Applications}, vol.~32, no.~6, p. 121, 2021.

\end{thebibliography}

\balance
\end{document}